\documentclass{article}

\PassOptionsToPackage{numbers, compress}{natbib}
\usepackage[final]{neurips_2019}

\usepackage[utf8]{inputenc} 
\usepackage[T1]{fontenc}    
\usepackage{hyperref}       
\usepackage{url}            
\usepackage{booktabs}       
\usepackage{amsfonts}       
\usepackage{nicefrac}       
\usepackage{microtype}      
\usepackage{amsmath}
\usepackage{graphicx}
\usepackage{float}
\usepackage{caption}
\usepackage{subcaption}
\usepackage{stackengine}
\usepackage{xspace}
\usepackage{enumitem}
\newcommand{\mindiff}{MinDiff\xspace}

\title{Toward a better trade-off between performance and fairness with kernel-based distribution matching}

\author{%
  Flavien Prost \\
  Google \\
  \texttt{fprost@google.com} \\
  \And
  Hai Qian \\
  Google \\
  \texttt{hqian@google.com} \\
  \And
  Qiuwen Chen \\
  Google \\
  \texttt{qiuwen@google.com} \\
  \And
  Ed H. Chi \\
  Google \\
  \texttt{edchi@google.com} \\
  \And
  Jilin Chen \\
  Google \\
  \texttt{jilinc@google.com} \\
  \And
  Alex Beutel \\
  Google \\
  \texttt{alexbeutel@google.com} \\
}

\begin{document}
\bibliographystyle{abbrvnat}
\maketitle

\vspace{-1mm}
\begin{abstract}
    As recent literature has demonstrated how classifiers often carry unintended biases toward some subgroups, deploying machine learned models to users demands careful consideration of the social consequences. 
    How should we address this problem in a real-world system? How should we balance core performance and fairness metrics?
    In this paper, we introduce a \mindiff framework for regularizing classifiers toward different fairness metrics and analyze a technique with kernel-based statistical dependency tests. 
    We run a thorough study on an academic dataset to compare the Pareto frontier achieved by different regularization approaches, and apply our kernel-based method to two large-scale industrial systems demonstrating real-world improvements.
\end{abstract}

\vspace{-2mm}
\section{Introduction}
\vspace{-2mm}
Over the last few years, the research community in machine learning has become more aware of the unintended biases that are learned and carried out by their models. Concerning behavior toward subgroups of the population has been pointed out in multiple applications ranging from the predictions of credit default \cite{HardtPS16} to the detection of abusive online comments \cite{Dixon}.

This growing collective awareness has resulted in significant interest to develop and popularize fairness metrics \cite{calders2010three, dwork2012fairness, HardtPS16, kearns2017preventing, Borkan, kallus2019fairness}, as well as to find efficient mitigation techniques \cite{Agarwal2018, Zafar, Gupta2016, Dixon,BeutelAdversarial, LiptonThreshold}. However these techniques typically come with multiple challenges as they might require collecting additional data (e.g., re-balancing techniques in \cite{Dixon}) or might generate instability in the training process (e.g., adversarial techniques in \cite{BeutelAdversarial, ZhangAdversarial, Madras}). 

We focus here on mitigation techniques in a classification setting and would like to ensure that our work is compatible with the challenges that real-world production systems face. We follow the work done by \cite{alexbeutelputting} in which the authors optimize for \emph{Equality of Opportunity} \cite{HardtPS16}, i.e. equalized false positive rate, by minimizing the correlation between the subgroup identity and the predictions over negative examples. Their method is shown to be efficient in a production system and was later generalized to a pairwise ranking loss for recommender settings \cite{alexbeutelrecommendation}.

Although this method gives good empirical results, the authors acknowledge that correlation does not guarantee statistical independence matching equality of opportunity. We therefore take inspiration from the work done by \citet{GrettonMMD}, which developed a kernel-based method to test the independence of two sample distributions, called Maximum Mean Discrepancy (MMD). This test has the advantage of being easy to compute and can be optimized by gradient-descent algorithms which made it useful for applications such as domain adaptation \cite{Long2015, Bousmalis2016}.

We focus on a mitigation technique which combines the MMD test and the framework of \cite{alexbeutelputting}. This approach was  mentioned in transfer learning for fairness \cite{Candice}, but has not previously been studied as a core bias mitigation technique; we analyze here its efficiency and practical application. Our main contributions are as follows:
\vspace{-1mm}
\begin{itemize}
	\item The definition of the \mindiff, a lightweight framework with a collection of regularization techniques for optimizing fairness metrics.
	\item An empirical comparison revealing the improvement of the performance/accuracy trade-off with kernel-based approaches.
	\item The description of the application of these techniques on two large scale production systems and the resulting improvements.
\end{itemize}

\section{\mindiff Framework}
\label{framework}

\vspace{-1mm}
\subsection{Setting and notations} 
\label{notation}

We consider a general task that consists in learning a function \(f\) that maps a set of features \(X\) to a binary label $Y \in \{0, 1\}$ such that \(P(y_{i}=1) = f(x_{i})\). We assume that the system makes an adversarial decision for each example where the predicted probability is above a certain threshold (undesirable outcome for the user). This setup is similar to many applications where the system might flag abusive comments \cite{Dixon} or reject an applicant's loan \cite{HardtPS16}.  
For each example, we can compute a loss \(L_{primary}(f(x_{i}), y_{i})\) (e.g. cross-entropy) and measure the primary performance of the model with traditional metrics such as accuracy.

We further assume that each example is associated to a subgroup \(A\), but this feature is available only for a fraction of the training data and is not observed at prediction time. For simplicity, we will focus on the binary case where \(A \in \{0, 1\}\) (out or in subgroup) however the concept applies to a more general setup.

We now want to evaluate the fairness of our model and consider the Equality of Opportunity \cite{HardtPS16} which is defined as an equality of the False Positive Rates across groups\footnote{Equality of opportunity can also be framed as equalized false negative rates.}. We measure the deviation from this ideal criteria with the False Positive Rate Gap (\(FPR_{gap}\)):
\vspace{-2mm}

\begin{equation*}
FPR_{gap} = | \Pr \{\hat{Y} = 1 | Y = 0, A = 1 \} - \Pr \{\hat{Y} = 1 | Y = 0, A = 0 \} |
\end{equation*}

\subsection{Mitigating bias with \mindiff}

Numerous regularization approaches have been proposed to encourage models during training to optimize for various fairness definitions \cite{Zafar, alexbeutelputting, Madras,Agarwal2018,ZhangAdversarial,Gupta2016}.
Here, we build off of the approach offered by \citet{alexbeutelputting}
to penalize the model for any dependence among the negative examples between the distribution of predicted probabilities and the subgroup label $A$. In practice, they add an additional regularization loss term to minimize the correlation between the two distributions. We give the naming ``\mindiff'' to this framework in relation to the fact that they minimize the difference of a given quantity between two slices of data.
\begin{equation}
\label{eqn:correlation}
Loss (X,Y) = L_{primary}(f(X), Y) + \lambda {\it Correlation} (f(X), A | Y = 0)
\end{equation}
where \(\lambda\) is a hyper parameter controlling the trade-off between primary and this \mindiff loss.

While correlation is not a sufficient test for statistical independence, the authors achieve good empirical results and claim that this is more easily adapted to real-world systems than other similar methods such as adversarial training \cite{BeutelAdversarial,Madras,ZhangAdversarial}.

We build on this technique and leverage the work done by \cite{GrettonMMD} who introduce a framework to test the statistical dependence between two sample distributions. Their statistic test, called Maximum Mean Discrepancy (MMD), consists in taking the mean between two samples  \(X_0\) and \(X_{1}\) mapped into a Reproducing Kernel Hilbert Subspace, and is easy to compute with the following formula:
\begin{equation*}
MMD (X_{0}, X_{1}) = \frac{1}{m^{2}} \sum_{i,j=1}^{m} k(x_{0,i}, x_{0,j}) -  \frac{2}{m n} \sum_{i,j=1}^{m,n} k(x_{0,i}, x_{1,j}) + \frac{1}{n^{2}} \sum_{i,j=1}^{n} k(x_{1,i}, x_{1,j})
\end{equation*}
where \((x_{0,i})_{i=1,m}\) are elements from the sample \(X_0\), \((x_{1,i})_{i=1,n}\) elements from \(X_1\) and  $k$ is a universal kernel (Gaussian or Laplace kernels are used in practice).

Based on this work, we suggest a new implementation of \mindiff. We add a similar loss component which penalizes any statistical dependence between the predictions for negative examples and the associated subgroup but we now measure it with the MMD function. Our new loss is then equal to:
\begin{equation}
\label{eqn:mmd}
Loss (X,Y) = L_{primary}(f(X), Y) + \lambda  {\it MMD} (f(X_{0}), f(X_{1})| Y = 0)
\end{equation}
where \(f(X_{0})\) are the predictions over examples where \(A = 0\), \(f(X_{1})\) are the predictions over examples where \(A = 1\), and \(\lambda\) is a hyperparameter controlling the trade-off between primary and \mindiff loss.

\vspace{-2mm}
\section{Academic Comparison}
\label{experience}

\newcommand{\fpr}{{\it FPR}}
\newcommand{\mmd}{{\it MMD}}

\vspace{-2mm}
\paragraph{Task description}
We use UCI's Adult dataset\footnote{\url{https://archive.ics.uci.edu/ml/datasets/adult}} \cite{UCIData} which contains census information over 40,000 individuals. The task is to predict if someone is earning more or less than \$50,000 (binary classification) and we use ``sex'' as the sensitive attribute, restricted in the dataset to binary values \{male, female\}. We use accuracy to measure the performance of the model and the \(\fpr_{gap}\) as the fairness metric.

We set the architecture to a feed-forward neural network with one hidden layer and selected the parameters with cross-validation: we used a learning rate of 0.001, a batch size of 256 examples and 64 hidden units. Without any bias mitigation, the model is 84.5\% accurate and has a 0.12 \(\fpr_{gap}\). 

We then train three different models by optimizing respectively the following losses: (a) The loss defined in equation \ref{eqn:correlation} (used in \cite{alexbeutelputting}), (b) Our new loss defined in equation \ref{eqn:mmd} with a Gaussian kernel, (c) Our new loss defined in equation \ref{eqn:mmd} with a Laplace kernel. We will use respectively the names \(Corr\), \(\mmd_{\it Gaussian}\) and \(\mmd_{\it Laplace}\) to refer to each of these models. 

Both Gaussian and Laplace required a parameter called \emph{kernel length} \(l\) (in the Gaussian kernel \(k(x,y) = \exp( - \frac{|| x - y ||^{2}}{l^{2}})\))  that we set to \( 0.1\) in order to be in the same order of magnitude as the standard deviation of the underlying distribution of probability. This value is motivated by the analysis reported in appendix \ref{ssec:parameter} where we describe how the value of this parameter impacts the results.

\begin{figure}
  \centering
  \includegraphics[width=0.9\linewidth]{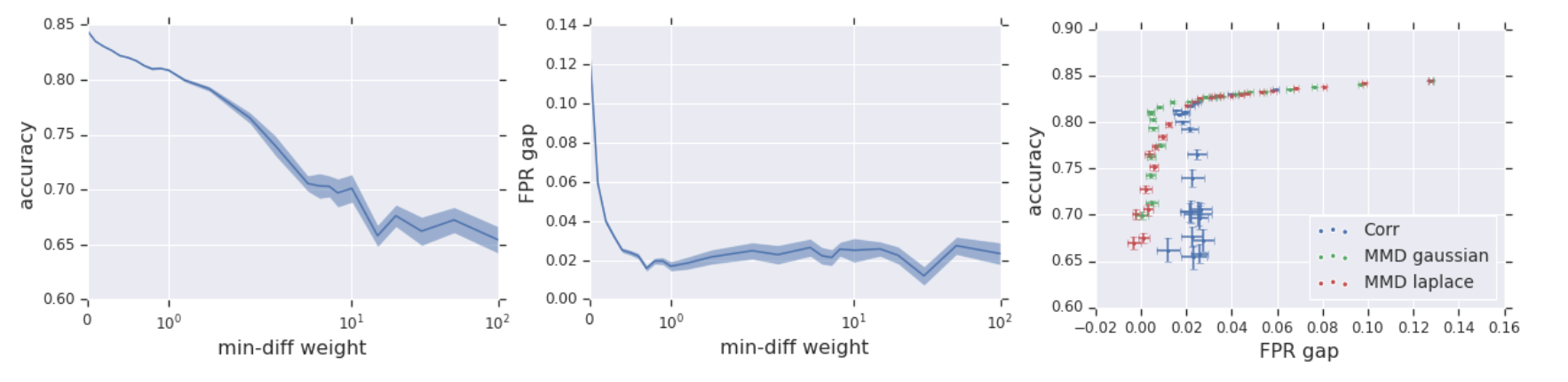}
  \caption{Visualization of the trade-off between performance and fairness. For the \(Corr\) model, as we increase the \mindiff weight, the accuracy gets worse (left) while the \(\fpr_{gap}\) decreases (middle). The right plot shows the Pareto frontier (accuracy vs \(\fpr_{gap}\)) for each \mindiff model.}
  \label{tradeoff}
  \vspace{-4mm}
\end{figure}

\vspace{-1mm}
\paragraph{Trade-off between accuracy and fairness}
First, we start by varying the weight associated to the \mindiff loss. For each value of \(\lambda\) we report the accuracy and \(\fpr_{gap}\) as the average values over twenty training runs. We also compute the standard error of the estimates. 

The resulting graphs for the \(Corr\) model can be found in the Figure \ref{tradeoff}. As expected, the \mindiff loss is able to reduce the \(\fpr_{gap}\) from 0.14 to 0.02 (left-plot), however this comes with a decrease in accuracy from 0.85 to 0.65 (middle-plot). The trade-off between those two metrics induced by the \(Corr\) model results in a Pareto-frontier plotted on the right.  

\vspace{-1mm}
\paragraph{MMD achieves a better ``Pareto frontier''}

The previous graphs help us to understand that these methods induce a significant trade-off between accuracy and fairness controlled by \(\lambda\) from Equation \eqref{eqn:mmd}. As a result, we want to analyze the variations of the Pareto-frontier with the three models and report it in Figure \ref{tradeoff} (right plot). We can see that, while both have similar effect with small values of \(\lambda\) (upper right part), \(\mmd\) outperforms \(Corr\) when we increase the fairness weight. In particular \(\mmd\) is able to reach low \(\fpr{gap}\) (< 0.01) with relatively high accuracy (>83\%) whereas the \(Corr\) model is unable to reduce the \(\fpr_{gap}\) below 0.02. We hypothesize that \(Corr\) is unable to remove entirely the bias as it only matches the mean and the variance of the two samples and not the true distribution.
Additionally, we observe that the choice of the kernel does not seem to have a large impact as \(\mmd_{\it Gaussian}\) and \(\mmd_{\it Laplace}\) have similar results.
\vspace{-2mm}

\section{Applications to real-world systems}
\label{applications}

\vspace{-1.5mm}
\subsection{Classifier Setting}
\vspace{-1mm}
The first system follows the same framework as the one introduced in \S\ref{notation}. We train a model $f$ with log-loss to predict a binary variable $Y$ for an item. When the predicted score is higher than a threshold $t$ (chosen for a fixed recall), we classify the item as ``positive'' (undesirable outcome).

\begin{table}
    \centering
    \begin{tabular}{c|c|c}
    \hline
         & Correlation & MMD\\
         \hline\hline
    FPR ratio     & 5.22 & 2.82\\
    \hline
    \end{tabular}
    \caption{The MMD loss results in significantly better FPR ratio than that of the correlation loss.}
    \label{tab:xyz}
    \vspace{-5mm}
\end{table}

The original analysis showed that the false positive rate (FPR) is higher for the protected group than for the majority group. Our goal is to minimize the ratio gap $FPR_{protected}/FPR_{majority}$.
Table \ref{tab:xyz} shows the comparison between different versions of \mindiff: MMD improves the fairness ratio gap over correlation loss by $45\%$, while both models maintain neutral performance for the main task.

\vspace{-2mm}
\subsection{Recommender System}
\vspace{-2mm}
We now focus on a large-scale, production recommender system where the model \(f\) is trained to predict the probability of an item being clicked (\(Y=1\)) by the user and is used in inference to score and rank the items to display. We consider a subgroup A of users as a variable in \(\{0, 1\}\).
We utilize the work done by \cite{alexbeutelrecommendation} who suggested a pairwise fairness metric for recommendation as well as a \mindiff formulation on pairs of items to improve the system that we summarize below.

\textbf{Fairness metric for ranking.}
For a random pair of items where exactly one of the items is clicked, we can define the \textit{pairwise ranking accuracy} as the frequency with which the model ranks higher the clicked item. With this, we can evaluate if the system is under-ranking a subgroup of items by computing the difference between the pairwise ranking accuracy when the clicked item is in or out of the subgroup (gap in pairwise ranking accuracy). We report this metric bucketed by level of satisfaction of the user after the click, as in \cite{alexbeutelrecommendation} and define the \emph{total gap} as the sum over all buckets.

\textbf{\mindiff formulation.}
To optimize the fairness metric, \cite{alexbeutelrecommendation} adds a \mindiff loss over random pairs of items \((x_{1}, x_{2})\) with (\(y_{1} = 1, y_{2} = 0\)) by penalizing the correlation between the following quantities:

\vbox{%
\begin{equation*}
    \alpha = f(x_{1}) - f(x_{2})\ \ = f(clicked) - f(unclicked)
\end{equation*}
\begin{equation*}
    \beta = A(x_{1}) - A(x_{2})\ \ = A(clicked) - A(unclicked)
\end{equation*}
}
\vspace{-2mm}
Our algorithm re-uses this formulation but penalizes the dependence between \(\alpha\) and \(\beta\) with \(MMD\).

\begin{figure}
    \centering
    \begin{subfigure}[b]{0.26\textwidth}
        \includegraphics[width=\textwidth]{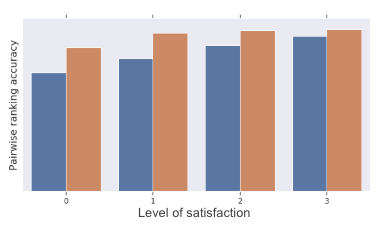}
        \caption{Original}
        \label{fig:original}
    \end{subfigure}%
    ~ 
    \begin{subfigure}[b]{0.26\textwidth}
        \includegraphics[width=\textwidth]{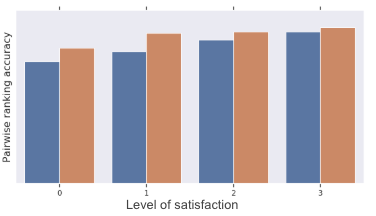}
        \caption{With \(Corr\)}
        \label{fig:corr}
    \end{subfigure}
    ~ 
    \begin{subfigure}[b]{0.26\textwidth}
        \includegraphics[width=\textwidth]{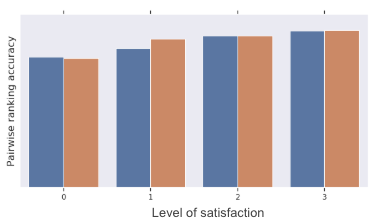}
        \caption{With MMD}
        \label{fig:mmd}
    \end{subfigure}
    \caption{Evolution of the gap in pairwise ranking accuracy}\label{fig:recresults}
    \vspace{-5mm}
\end{figure}

\textbf{Results of our algorithm.} We compare our MMD approach to the initial system and to the \(Corr\) method of \cite{alexbeutelrecommendation} and display the results in Figure \ref{fig:recresults}. While the \mindiff approach with correlation reduced the total gap in pairwise ranking accuracy by 60\%, MMD is able to bring the gap even lower (reduction of 65\% from the \(Corr\) number). Additionally, online experiments showed that these features came with neutral impact on overall system performance.

\vspace{-3mm}
\section{Conclusion}
\vspace{-2mm}

In this paper, we define the \mindiff framework as a collection of regularization techniques for mitigating bias and analyze approaches based on the MMD statistical test. We show empirically that they achieve a better Pareto-frontier and describe two applications of it to real-world systems. With this work, we hope to help reduce the challenges in bringing ML Fairness to industry applications.

\medskip

\bibliography{references}

\begin{thebibliography}{21}
\providecommand{\natexlab}[1]{#1}
\providecommand{\url}[1]{\texttt{#1}}
\expandafter\ifx\csname urlstyle\endcsname\relax
  \providecommand{\doi}[1]{doi: #1}\else
  \providecommand{\doi}{doi: \begingroup \urlstyle{rm}\Url}\fi

\bibitem[Agarwal et~al.(2018)Agarwal, Beygelzimer, Dud{\'{\i}}k, Langford, and
  Wallach]{Agarwal2018}
A.~Agarwal, A.~Beygelzimer, M.~Dud{\'{\i}}k, J.~Langford, and H.~M. Wallach.
\newblock A reductions approach to fair classification.
\newblock \emph{CoRR}, abs/1803.02453, 2018.
\newblock URL \url{http://arxiv.org/abs/1803.02453}.

\bibitem[Beutel et~al.(2017)Beutel, Chen, Zhao, and Chi]{BeutelAdversarial}
A.~Beutel, J.~Chen, Z.~Zhao, and E.~H. Chi.
\newblock Data decisions and theoretical implications when adversarially
  learning fair representations.
\newblock \emph{CoRR}, abs/1707.00075, 2017.
\newblock URL \url{http://arxiv.org/abs/1707.00075}.

\bibitem[Beutel et~al.(2019{\natexlab{a}})Beutel, Chen, Doshi, Qian, Wei, Wu,
  Heldt, Zhao, Hong, Chi, and Goodrow]{alexbeutelrecommendation}
A.~Beutel, J.~Chen, T.~Doshi, H.~Qian, L.~Wei, Y.~Wu, L.~Heldt, Z.~Zhao,
  L.~Hong, E.~H. Chi, and C.~Goodrow.
\newblock Fairness in recommendation ranking through pairwise comparisons.
\newblock \emph{CoRR}, abs/1903.00780, 2019{\natexlab{a}}.
\newblock URL \url{http://arxiv.org/abs/1903.00780}.

\bibitem[Beutel et~al.(2019{\natexlab{b}})Beutel, Chen, Doshi, Qian, Woodruff,
  Luu, Kreitmann, Bischof, and Chi]{alexbeutelputting}
A.~Beutel, J.~Chen, T.~Doshi, H.~Qian, A.~Woodruff, C.~Luu, P.~Kreitmann,
  J.~Bischof, and E.~H. Chi.
\newblock Putting fairness principles into practice: Challenges, metrics, and
  improvements.
\newblock \emph{CoRR}, abs/1901.04562, 2019{\natexlab{b}}.
\newblock URL \url{http://arxiv.org/abs/1901.04562}.

\bibitem[Borkan et~al.(2019)Borkan, Dixon, Sorensen, Thain, and
  Vasserman]{Borkan}
D.~Borkan, L.~Dixon, J.~Sorensen, N.~Thain, and L.~Vasserman.
\newblock Nuanced metrics for measuring unintended bias with real data for text
  classification.
\newblock \emph{CoRR}, abs/1903.04561, 2019.
\newblock URL \url{http://arxiv.org/abs/1903.04561}.

\bibitem[Bousmalis et~al.(2016)Bousmalis, Trigeorgis, Silberman, Krishnan, and
  Erhan]{Bousmalis2016}
K.~Bousmalis, G.~Trigeorgis, N.~Silberman, D.~Krishnan, and D.~Erhan.
\newblock Domain separation networks.
\newblock \emph{CoRR}, abs/1608.06019, 2016.
\newblock URL \url{http://arxiv.org/abs/1608.06019}.

\bibitem[Calders and Verwer(2010)]{calders2010three}
T.~Calders and S.~Verwer.
\newblock Three naive bayes approaches for discrimination-free classification.
\newblock \emph{Data Mining and Knowledge Discovery}, 21\penalty0 (2):\penalty0
  277--292, 2010.

\bibitem[Dixon et~al.(2018)Dixon, Li, Sorensen, Thain, and Vasserman]{Dixon}
L.~Dixon, J.~Li, J.~Sorensen, N.~Thain, and L.~Vasserman.
\newblock Measuring and mitigating unintended bias in text classification.
\newblock In \emph{Proceedings of the 2018 AAAI/ACM Conference on AI, Ethics,
  and Society}, AIES '18, pages 67--73, New York, NY, USA, 2018. ACM.
\newblock ISBN 978-1-4503-6012-8.
\newblock \doi{10.1145/3278721.3278729}.
\newblock URL \url{http://doi.acm.org/10.1145/3278721.3278729}.

\bibitem[Dua and Graff(2017)]{UCIData}
D.~Dua and C.~Graff.
\newblock {UCI} machine learning repository, 2017.
\newblock URL \url{http://archive.ics.uci.edu/ml}.

\bibitem[Dwork et~al.(2012)Dwork, Hardt, Pitassi, Reingold, and
  Zemel]{dwork2012fairness}
C.~Dwork, M.~Hardt, T.~Pitassi, O.~Reingold, and R.~Zemel.
\newblock Fairness through awareness.
\newblock In \emph{Proceedings of the 3rd innovations in theoretical computer
  science conference}, pages 214--226. ACM, 2012.

\bibitem[Goh et~al.(2016)Goh, Cotter, Gupta, and Friedlander]{Gupta2016}
G.~Goh, A.~Cotter, M.~Gupta, and M.~P. Friedlander.
\newblock Satisfying real-world goals with dataset constraints.
\newblock In D.~D. Lee, M.~Sugiyama, U.~V. Luxburg, I.~Guyon, and R.~Garnett,
  editors, \emph{Advances in Neural Information Processing Systems 29}, pages
  2415--2423. Curran Associates, Inc., 2016.
\newblock URL
  \url{http://papers.nips.cc/paper/6316-satisfying-real-world-goals-with-dataset-constraints.pdf}.

\bibitem[Gretton et~al.(2008)Gretton, Borgwardt, Rasch, Sch{\"{o}}lkopf, and
  Smola]{GrettonMMD}
A.~Gretton, K.~M. Borgwardt, M.~J. Rasch, B.~Sch{\"{o}}lkopf, and A.~J. Smola.
\newblock A kernel method for the two-sample problem.
\newblock \emph{CoRR}, abs/0805.2368, 2008.
\newblock URL \url{http://arxiv.org/abs/0805.2368}.

\bibitem[Hardt et~al.(2016)Hardt, Price, and Srebro]{HardtPS16}
M.~Hardt, E.~Price, and N.~Srebro.
\newblock Equality of opportunity in supervised learning.
\newblock \emph{CoRR}, abs/1610.02413, 2016.
\newblock URL \url{http://arxiv.org/abs/1610.02413}.

\bibitem[Kallus and Zhou(2019)]{kallus2019fairness}
N.~Kallus and A.~Zhou.
\newblock The fairness of risk scores beyond classification: Bipartite ranking
  and the xauc metric.
\newblock \emph{arXiv preprint arXiv:1902.05826}, 2019.

\bibitem[Kearns et~al.(2017)Kearns, Neel, Roth, and Wu]{kearns2017preventing}
M.~Kearns, S.~Neel, A.~Roth, and Z.~S. Wu.
\newblock Preventing fairness gerrymandering: Auditing and learning for
  subgroup fairness.
\newblock \emph{arXiv preprint arXiv:1711.05144}, 2017.

\bibitem[Lipton et~al.(2018)Lipton, McAuley, and Chouldechova]{LiptonThreshold}
Z.~Lipton, J.~McAuley, and A.~Chouldechova.
\newblock Does mitigating ml's impact disparity require treatment disparity?
\newblock In S.~Bengio, H.~Wallach, H.~Larochelle, K.~Grauman, N.~Cesa-Bianchi,
  and R.~Garnett, editors, \emph{Advances in Neural Information Processing
  Systems 31}, pages 8125--8135. Curran Associates, Inc., 2018.
\newblock URL
  \url{http://papers.nips.cc/paper/8035-does-mitigating-mls-impact-disparity-require-treatment-disparity.pdf}.

\bibitem[Long et~al.(2015)Long, Cao, Wang, and Jordan]{Long2015}
M.~Long, Y.~Cao, J.~Wang, and M.~I. Jordan.
\newblock Learning transferable features with deep adaptation networks.
\newblock In \emph{Proceedings of the 32Nd International Conference on
  International Conference on Machine Learning - Volume 37}, ICML'15, pages
  97--105. JMLR.org, 2015.
\newblock URL \url{http://dl.acm.org/citation.cfm?id=3045118.3045130}.

\bibitem[Madras et~al.(2018)Madras, Creager, Pitassi, and Zemel]{Madras}
D.~Madras, E.~Creager, T.~Pitassi, and R.~S. Zemel.
\newblock Learning adversarially fair and transferable representations.
\newblock \emph{CoRR}, abs/1802.06309, 2018.
\newblock URL \url{http://arxiv.org/abs/1802.06309}.

\bibitem[Schumann et~al.(2019)Schumann, Wang, Beutel, Chen, Qian, and
  Chi]{Candice}
C.~Schumann, X.~Wang, A.~Beutel, J.~Chen, H.~Qian, and E.~H. Chi.
\newblock Transfer of machine learning fairness across domains.
\newblock \emph{CoRR}, abs/1906.09688, 2019.
\newblock URL \url{http://arxiv.org/abs/1906.09688}.

\bibitem[Zafar et~al.(2015)Zafar, Valera, Rodriguez, and Gummadi]{Zafar}
M.~Zafar, I.~Valera, M.~Rodriguez, and K.~P. Gummadi.
\newblock Fairness constraints: A mechanism for fair classification.
\newblock 07 2015.

\bibitem[Zhang et~al.(2018)Zhang, Lemoine, and Mitchell]{ZhangAdversarial}
B.~H. Zhang, B.~Lemoine, and M.~Mitchell.
\newblock Mitigating unwanted biases with adversarial learning.
\newblock \emph{CoRR}, abs/1801.07593, 2018.
\newblock URL \url{http://arxiv.org/abs/1801.07593}.

\end{thebibliography}

\newpage
 
\appendix
\section*{Appendices}
\addcontentsline{toc}{section}{Appendices}
\renewcommand{\thesubsection}{\Alph{subsection}}

\subsection{Heuristics for values of the kernel length in the  Gaussian and Laplace kernels} \label{ssec:parameter}

We use the following formulas to compute the Gaussian and Laplace kernels:

\begin{equation*}
    GaussianK (x,y) = \exp( - \frac{|| x - y ||^{2}}{l^{2}})
\end{equation*}
\begin{equation*}
    LaplaceK (x,y) = \exp( - \frac{|| x - y ||}{l})
\end{equation*}
where \(l\) is a parameter called kernel length.

We want to analyze how the parameter \(l\) impacts previous results and what value is optimal. We describe here our analysis only for the Gaussian kernel even if similar results apply to the Laplace one. 

\begin{figure}[H]
  \centering
  \includegraphics[width=\linewidth]{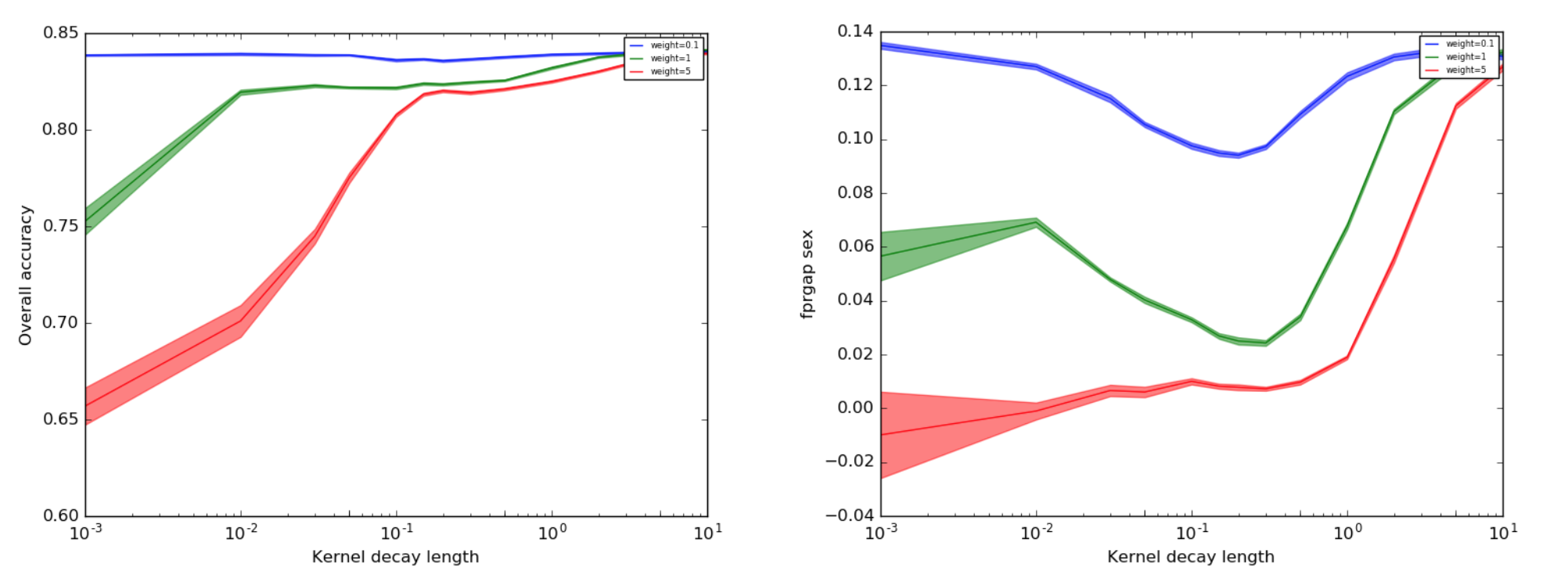}
  \caption{Impact of the kernel decay length on the \(MMD_{Gaussian}\) model}
  \label{kernellength}
\end{figure}

We vary the value of \(l\) for three values of \(\lambda\) (0.1, 1 and 5) and report the results in Figure \ref{kernellength}. As metrics are not monotonous functions of \(l\), the Pareto frontier is harder to read and we therefore do not report it in this paper.

We observe that, if the accuracy decreases slightly between \(10^{1}\) and \(10^{-1}\) and then drops as we decrease \(l\) further (left plot). We believe that this is due to the fact that smaller values of \(l\) make the kernel more sensitive, as a result, the \(MMD\) returns higher values for any differences between the distributions. On the other side, \(FPR_{gap}\) has a more complicated relationship and seems to be optimal for intermediate values of kernel length (right plot).

Overall, there seems to be a sweet spot around 0.1 and 0.5 where we reach a good trade-off between fairness and accuracy. We observe that this range is close to the standard deviation of the underlying probability distribution and hypothesize that this heuristic should generalize to any distributions.

\end{document}